\documentclass[runningheads, envcountsame, a4paper]{llncs}
\usepackage{graphicx}
\usepackage[misc]{ifsym}
\usepackage{subcaption}
\usepackage{algorithm,algorithmic}
\usepackage{amsmath}
\usepackage{makecell}
\usepackage{multirow}
\usepackage{array}
\usepackage{microtype}
\begin{document}
\title{TagRec: Automated Tagging of Questions with Hierarchical Learning Taxonomy \thanks{This work is supported by Extramarks Education (an education technology company) and TiH Anubhuti (IIITD).}}
\titlerunning{Automated Tagging of Questions with Hierarchical Learning Taxonomy}
%
%
\author{Venktesh V \and
Mukesh Mohania \and
Vikram Goyal}
%
\authorrunning{Venktesh V. et al.}
%
\institute{Indraprastha Institute of Information Technology, Delhi \email{\{venkteshv,mukesh,vikram\}@iiitd.ac.in}
}

\maketitle   \setcounter{footnote}{0} 
\begin{abstract}
Online educational platforms organize academic questions based on a hierarchical learning taxonomy (subject-chapter-topic). Automatically tagging new questions with existing taxonomy will help organize these questions into different classes of hierarchical taxonomy so that they can be searched based on the facets like chapter, topic. This task can be formulated as a flat multi-class classification problem. Usually, flat classification based methods ignore the semantic relatedness between the terms in the hierarchical taxonomy and the questions. Some traditional methods also suffer from the class imbalance issues as they consider only the leaf nodes ignoring the hierarchy.  Hence, we formulate the problem as a similarity-based retrieval task where we optimize the semantic relatedness between the taxonomy and the questions. We demonstrate that our method helps to handle the unseen labels and hence can be used for taxonomy tagging in the wild, like the question-answer forums. In this method, we augment the question with its corresponding answer to capture more semantic information and then align the question-answer pair's contextualized embedding with the corresponding label (taxonomy) vector representations. The representations are aligned by fine-tuning a transformer based model with a loss function that is a combination of the cosine similarity and hinge rank loss. The loss function maximizes the similarity between the question-answer pair and the correct label representations and minimizes the similarity to unrelated labels. Finally, we perform extensive experiments on two real-world datasets. We empirically show that the proposed learning method outperforms representations learned using the multi-class classification method and other state of the art methods by \textbf{6\%} as measured by Recall@k. We also demonstrate the performance of the proposed method on unseen but related learning content like the learning objectives without re-training the network.
\end{abstract}
\begin{figure*}
\captionsetup{justification=centering}

\begin{subfigure}[t]{0.59\linewidth}
\includegraphics[width=1\linewidth]{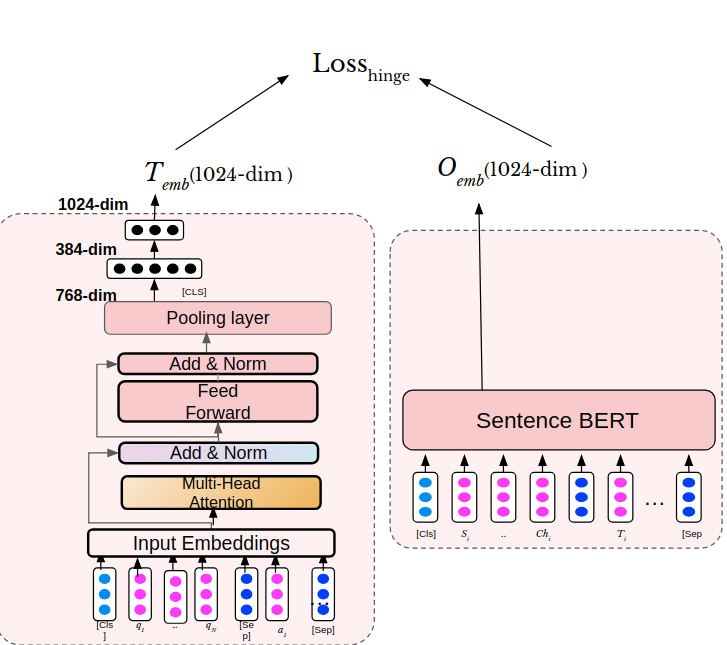}
\caption{Training phase - aligns \\input and label embeddings. }
\end{subfigure}%
\hspace{2em}
\begin{subfigure}[t]{0.48\linewidth}
\includegraphics[width=0.9\linewidth]{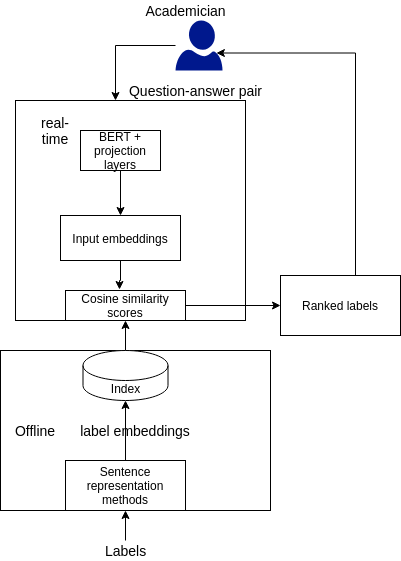}
\caption{Testing (inference) phase - \\ recommends labels}
\end{subfigure}%
\caption{Training and testing phases for tagging questions with hierarchical labels}
\end{figure*}
\section{Introduction}
Online learning platforms organize academic questions according to a hierarchical learning taxonomy (subject-chapter-topic). For instance a question about "electromotive force"  is tagged with \textbf{"science - physics - electricity"}. This method of organization helps individuals navigate over large question banks. The taxonomy can also aid in \textit{faceted} search. The \textit{facets} could be topics, concepts, or chapters. However, manually tagging each question with the appropriate learning taxonomy is cumbersome. Hence there is a need for automated methods for tagging a question with the appropriate learning taxonomy. Automated tagging helps to organize acquired questions from third party vendors, which may be rarely linked to a learning taxonomy or are linked only at a ”chapter” level. Also, the learning taxonomy is subject to change as the topic names or concept names could be replaced by synonyms or related concepts. Hence, the taxonomy tagging method should adapt to minor changes in the label (taxonomy) space without changes in the model architecture or re-training.


Automated categorization of content in online platforms is usually formulated as a multi-class classification problem \cite{xumulti,kozareva2015everyone}. However, there are some unique challenges when dealing with a hierarchical taxonomy and tagging short questions in the e-learning domain. \textit{Firstly}, some of the traditional multi-class classification methods ignore the hierarchy and consider only leaf nodes of the hierarchical labels as labels. However, this formulation of the problem would suffer from class imbalance issues since a large number of contents may be tagged with a small number of leaf nodes leaving a smaller number of samples for other leaf nodes. The \textit{second challenge} is that the labels are dynamic in nature as new topics could be added to the syllabus, and the old topics may no longer be valid or could be retired. This results in a change in the label space and thus gives rise to new labels. The new labels would have some similarity to some of the existing labels as the subject name and the chapter names could be semantically related to the existing chapter names. The traditional multi-class classification methods cannot exploit this semantic relatedness as they do not consider label representations. They require a change in architecture to incorporate the new labels and must be retrained. However, the hierarchical labels are an abstraction of their word descriptions and hence some of the terms in the hierarchical labels are semantically related to the words in the given questions. Hence, by learning a representation that captures the similarity between the labels and the related questions, the model can adapt to changes in label space.

To capture more semantic information from the given inputs, we augment the question with its answer as an auxiliary information. Hence, we refer to the augmented content as a \textit{"question-answer"} pair and the hierarchical learning taxonomy is referred to as \textit{"label"} or \textit{"taxonomy"}. Our method, however would work even in cases where the answer is not given along with the question.

We propose a new method, named \textbf{TagRec}, for question-answer categorization in online learning platforms. In our method, the goal is to recommend relevant hierarchical learning taxonomy (label) for every question-answer pair to assist in organizing the learning content. Hence we adopt a similarity based retrieval method where hierarchical labels which are semantically related to the given question-answer pair. Figure 1 shows the basic architecture of the proposed method. Here, in the Figure 1(a), the method projects the  question-answer text and the corresponding label as inputs to a continuous vector space and aligns the input representations $T_{emb}$ with the label representations $O_{emb}$. In the Figure 1(b), during the recommendation (test time), when a new question arrives, the method projects the new question-answer pair to the vector space and computes the cosine similarity between the input representations and vector representations of all known labels. The labels are then ranked according to the similarity score, and the top-k labels are recommended for the given new question.

The proposed method can be used for tag recommendation in open source platforms like \textit{StackExchange}. For example, a question about "Batch normalization" with tags "deep-learning" and "normalization" can be tagged with a hierarchical label \textit{AI$\xrightarrow{}$deep learning$\xrightarrow{}$normalization$\xrightarrow{}$Batch normalization}.  The preprocessed data can then be fed to TagRec, which would be able to recommend hierarchical labels to new questions after the training.


The following are the key technical contributions of the paper:
\let\labelitemi\labelitemii
\begin{itemize}
 \setlength\itemsep{0.1em}

    \item We propose a novel and efficient similarity based retrieval method to recommend a hierarchical taxonomy label to a given question-answer pair. The method decouples the computation of vector representations for the question input and the taxonomy labels, thus allowing label representations to be pre-computed and indexed for lookup.
    \item We propose a learning method to align the input and hierarchical label representations that involves a loss function combining the cosine similarity and the hinge rank loss \cite{frome2013devise}.
    \item  We employ a transformer based sentence representation method to represent the hierarchical labels. We conduct extensive experiments by varying the label representations in the architecture shown in Figure 1(a) to empirically determine the effect of the label representations on the performance of the method. The proposed TagRec method outperforms the state of the art methods by upto \textbf{6\%} with Recall@k as the metric.
    \item We demonstrate the ability of our method to adapt the changes in label space without any changes in architecture or retraining. 
     \item We further demonstrate the ability of our method to categorize the unseen but related learning content like learning objectives. We extract 417 learning objectives from science textbooks and apply the proposed method to this data without any re-training. We observe that the proposed method is able to achieve high Recall@k at top-2 predictions and outperforms the existing state of the art methods by \textbf{7\%}.
\end{itemize}

\section{Related Work}

In this section, we first provide an overview of multi-class classification methods that consider the hierarchical label structure and then briefly discuss the current state of the art sentence representation methods.

\setlength{\parindent}{0pt}

\subsection{Multi-class classification with hierarchical taxonomy}
\setlength{\parindent}{0pt}
Many websites in the e-commerce and e-learning domains organize their content based on a hierarchical taxonomy \cite{xumulti,kozareva2015everyone}. The most common approaches for automatic categorization of the content to the hierarchical labels are flat multi-class single-step classification  and hierarchical multi-step classifiers \cite{article,yu2012product}. In multi-class single-step methods, the hierarchy is ignored and the leaf nodes are considered as labels. This leads to class imbalance issue, as discussed in Section 1. In the hierarchical multi-step approach, a classifier is trained to predict the top-level category and the process is repeated for predicting the sub-categories. However, the main problems associated with this approach are that the error from the classifiers at one level propagates to the next level and the number of classifiers increases at every step.

Several single-step classifiers have been proposed for the task of hierarchical classification. In \cite{yu2012product}, the word level features like n-grams were used with SVM as classifier to predict level 1 categories, whereas in \cite{kozareva2015everyone} the authors have leveraged n-gram features and distributed representations from Word2Vec to obtain features and fed them to a linear classifier for multi-class classification. Several deep learning methods like CNN \cite{10.1145/3302425.3302483} and LSTM \cite{article} have been proposed for the task of question classification. Since the pre-trained language models, like BERT \cite{BERT}, improve the performance, the authors in \cite{xumulti} propose a model BERT-QC, which fine tunes BERT on a sample of questions from science domain to classify them to a hierarchical taxonomy. The hierarchical multi-class classification problem has also been cast as a machine translation problem in \cite{MachinT} where the authors provide the product titles as input and use a seq2seq architecture to translate them to product categories that exhibit a hierarchy. However, all these above approaches do not consider the label representations. The hierarchical neural attention model \cite{sinha2018hierarchical} has been proposed, which leverages attention to obtain useful input sentence representation and uses an encoder-decoder architecture to predict each category in the hierarchical taxonomy. However, this approach may not scale to deep hierarchies.

\setlength{\parindent}{4ex}In this paper, we take a similarity-based retrieval approach with the aim to recommend the relevant label (i.e., the hierarchical learning taxonomy) by aligning the input embeddings and the label embeddings. We do not explore the multi-level classifier approach owing to the shortcomings explained earlier in this section. The proposed method can also adapt to changes in the label space.

\subsection{Sentence representation methods}
Distributed representations that capture the semantic relationships \cite{10.5555/2999792.2999959} have helped to advance many NLP tasks like classification, retrieval. Methods like GloVe \cite{pennington-etal-2014-glove} learn vector representation of word by performing dimensionality reduction on a co-occurrence count matrix. Rather than averaging word representations to obtain sentence embeddings, an unsupervised method named Sent2Vec \cite{pagliardini2017unsupervised} for composing n-gram embeddings to learn sentence representations was proposed.

The Bidirectional Encoder Representation from Transformers (BERT) \cite{BERT} is one of the current state of the art methods. However, one of the disadvantages of the BERT network structure is that no independent sentence embeddings are computed.
The Sentence-BERT \cite{reimers-gurevych-2019-sentence} model was proposed to generate useful sentence embeddings by fine-tuning BERT. Another transformer based sentence encoding model is the Universal Sentence Encoder (USE) \cite{cer-etal-2018-universal}  that has been specifically trained on semantic textual similarity task and generates useful sentence representations.

\setlength{\parindent}{4ex}In this paper, we treat each label as a sentence and embed it using the sentence representation methods. For example, the label \textbf{Science - Physics - electricity} is treated as a sentence. In our experiments, we observe that USE embeddings and Sentence-BERT embeddings perform better than averaging word embeddings. These results are discussed in Section 4.
 \begin{algorithm}
 \caption{Tag Recommender}
 \label{algoBO}
 \begin{algorithmic}[1]
 \renewcommand{\algorithmicrequire}{\textbf{Input:}}
 \renewcommand{\algorithmicensure}{\textbf{Output:}}
 \REQUIRE Training set $T \gets $ docs $\{D_1,..D_n\}$, labels $O$ of form (Subject-Chapter-Topic)\\
 \ENSURE  Set of tags for test set , $RO$
  \\ \textbf{Training} (batch mode)
  \STATE Get input text embeddings , $T_{emb} \gets BERT(D)$
  \STATE Obtain label embeddings, $O_{emb} \gets SENT\_BERT(O)$ 
  \STATE $Index(labels) \gets O_{emb}$
    \STATE  $loss\gets$ $\sum_{j \neq label}max(0,margin-cos(T_{emb},O_{emb}(label))+cos(T_{emb},O_{emb}(j)))$
    \STATE Fine-tune BERT to minimize $loss$ and align $T_{emb}$ and $O_{emb}$
 \\ \textbf{Testing Phase}
 \STATE Compute embeddings for test set $S$ using fine-tuned BERT $S_{emb} \gets BERT(S)$
 \STATE Rank set of unique labels $RO \gets sorted(Sim(S_{emb},O_{emb}))$
 \RETURN Top-k labels from $RO$
 \end{algorithmic} 
 \end{algorithm}
\section{Methodology}
In this section, we describe our method for classifying questions to  hierarchical labels. The method consists of a training phase and testing phase, as shown in Figure 1. The input to the method is a corpus of documents, $C=\{D_1,D_2...D_n\}$ where each document corresponds to a question-answer pair and the hierarchical labels $O=\{(S_1,Ch_1,T_1),(S_2,Ch_2,T_2)...\}$ where $S_i$,
$Ch_i$ and $T_i$ denote subject, chapter, and topic respectively. The goal here is to learn an input representation that is close to the correct label in the vector space. We consider the label $(S_i,Ch_i,T_i)$ as a sequence, ($S_i+Ch_i+T_i$) and obtain a sentence representation for it using pre-trained models. We obtain contextualized representations for the inputs using BERT \cite{BERT} followed by two projection layers. The linear projection layers are transformations that map the 768-D representation from BERT to the 1024-D or 512-D vector representation.

The steps of the proposed method are given in Algorithm 1.
The details of the two phases in Algorithm 1 are as follows:
 \setlength{\parindent}{0pt}
\let\labelitemi\labelitemii
\begin{itemize}
  \setlength\itemsep{0.5em}
\item In the \textit{training} phase, the input question-answer pair is passed through a transformer based language model BERT followed by projection layers. The vector representations for the labels are obtained using a sentence representation method like USE \cite{cer-etal-2018-universal} or Sentence-BERT \cite{reimers-gurevych-2019-sentence}. The vector representations for all unique set of labels can be pre-computed and indexed for lookup. This saves computation cost and time during training and testing phases. The model is fine-tuned using a loss function that is a combination of cosine similarity and hinge rank loss \cite{frome2013devise}. This helps to align the contextualized input representations with the label representations.

\item In the \textit{testing} phase, as shown in Figure 1b, the results are obtained in three steps. Firstly, the vector representations (embedding) for the input are computed using the fine-tuned BERT model. Secondly, the labels are ranked by computing cosine similarity between the input embeddings and the pre-computed label embeddings. Finally, top-k labels are chosen and metrics like Recall@k are computed for evaluating the performance of the model.
    \end{itemize}

\setlength{\parindent}{4ex}Our method is efficient as the label representations are pre-computed and indexed. Hence the time complexity at inference or testing time is $O(T_{M}N_{qa})$, where $T_{M}$ is the time cost of the model (BERT + projection layers) and $N_{qa}$ is the number of question-answer pairs at test time.
\subsection{Contextualized Input representations}
The academic questions are mostly comprised of technical terms or concepts that are related with the "topic" component of the label. For example, a question that contains terms like \textit{"ethyl alcohol"} is closely related with the topic \textit{"alcohols and ethers"} and hence the question can be tagged with the label \textit{"science - chemistry - alcohols and ethers"}. Academic questions also have terms that refer to different meanings depending on the context of their occurrence in the input sentence. For instance, the word "imaginary" in the sentence "Consider an imaginary situation" and its occurrence in the sentence "Given two imaginary numbers" has different meanings. This is an example of \textbf{polysemy} where the same word has different meanings in different contexts. Hence we need a method that can focus  on important terms in the sequence and also tackle the problem of polysemy. To tackle the mentioned problems, we use a transformer based language model BERT for projecting the input text to the vector space. The BERT is a language model where the representations are learnt in two stages. In the first stage, the model is trained in an unsupervised manner. In the second stage, the model is fine-tuned on task specific labelled data to produce representations for downstream tasks. The "self-attention" mechanism in BERT helps in obtaining better vector representations and helps tackle the problem of polysemy.

Self-attention \cite{vaswani2017attention} is the core of transformer based language models, and BERT leverages it to obtain better representation for a word by attending other relevant words in the context; Thus, a word has different representations depending on the context it has been used in. Self-attention encodes each word in the sentence using Query (Q), Key(K) and Value(V) vectors to obtain attention scores which determines how much attention to pay to each word when generating an embedding for the current word. Mathematically, 
\begin{eqnarray}
        Attention(Q,K,V) & = & \dfrac{Softmax(Q * K^T)}{\sqrt{d_k}} *V \\
        Softmax(x_i) & = & \dfrac{exp(x_i)}{\sum_j^N exp(x_j)} 
\end{eqnarray}

where $d_k$ is the dimension of query, key, and value vectors and is used to scale the attention scores. 

The self-attention mechanism helps to obtain contextualized representations that tackle the mentioned problems. We obtain contextualized representations of the input from BERT and pass them through the two projection layers, as shown in Figure 1a. We fine-tune BERT and the projection layers to align the generated contextualized representations with label representations as given in Algorithm 1. We further explore the training phase in Section 3.3

\subsection{Hierarchical label representations}
Here, we describe how sentence representations are obtained for the labels. We consider the labels that have a hierarchical structure as a sequence of words and leverage sentence embedding methods to project them to vector space. We embed the labels this way to preserve the semantic relatedness between the labels. For instance, the label like \textbf{science - physics - electricity} must be closer to \textbf{science - physics - magnetism} than \textbf{science - biology - biomolecules} in the vector space. With simple vector arithmetic (cosine similarity), we observe that embedding the labels with sentence based representation methods like Sentence-BERT or Sent2Vec help to preserve the semantic relatedness when compared to averaging word embeddings from GLoVe \cite{pennington-etal-2014-glove}. The sentence representation methods also do not suffer from constituent words being out of vocabulary unlike traditional word embedding methods and are able to handle such words. Since the Sentence-BERT and the USE models have been explicitly trained on semantic textual similarity tasks they provide rich textual representations that can be used for similarity based retrieval tasks. Hence, in this paper, we extensively experiment with various sentence embeddings methods like Sent2Vec, Universal Sentence Encoder (USE), and Sentence-BERT. We also propose a method where the labels are represented using the mean of the  GloVe vectors. We observe that sentence embedding methods significantly outperform the averaging of word vectors. The results are discussed in detail in the \textbf{Experiments and Results} section.

\subsection{Loss function}
In the training phase in Algorithm 1,  hinge rank loss is employed to maximize the similarity between contextualized input text embeddings and the vector representation of the correct label.

The hinge ranking loss is defined as :
 \[loss(text,label)\gets \sum_{j \neq label}max(0,margin-cos(T_{emb},v(label))+cos(T_{emb},v(j)))\]
 where $T_{emb}$ denotes the input text embeddings from BERT,
$v(label)$ denotes the vector representation of the correct label, $v(j)$ denotes the vector representation of an incorrect label, and $cos$ denotes the cosine similarity function. The derivative of the loss function is propagated, and the linear projection layers are trained and the BERT layers are fine-tuned to minimize the loss as given in Algorithm 1. The margin was set to a value of 0.1, which is a fraction of the norm of the embedding vectors (1.0), and it yields the best performance.
   \begin{table}
  \small
 \centering
\caption{Some samples from the QC-Science dataset}\label{tab1}
\begin{tabular}{|p{3.2cm}|p{4cm}|p{4.8cm}|}
\hline
Question &  Answer & Taxonomy\\
\hline
The value of electron gain enthalpy of chlorine is more than that of fluorine.  Give reasons &  Fluorine atom is small  so electron charge density on F atom is very high & Science$\xrightarrow{}$chemistry$\xrightarrow{}$classification of elements and periodicity in properties\\ \hline
What are artificial sweetening agents? & The chemical substances which are sweet in taste but do not add any calorie & Science$\xrightarrow{}$chemistry$\xrightarrow{}$chemistry in everyday life\\
\hline
\end{tabular}
\end{table}

 \section{Experiments}
 In this section, we discuss the experimental setup and the datasets on which the experiments were performed. All experiments are carried out on Google colab.

 \subsection{Datasets}
 To evaluate the effectiveness of the proposed method, we perform experiments on the following datasets:
 \let\labelitemi\labelitemii
 \begin{itemize}
     \item \textbf{QC-Science}: This dataset contains 47832 question-answer pairs belonging to the science domain tagged with labels of the form subject - chapter - topic. The dataset was collected with the help of a leading e-learning platform. The dataset consists of 40895 samples for training, 2153 samples for validation and 4784 samples for testing. Some samples are shown in Table 1. The average number of words per question is 37.14, and per answer, it is 32.01.
     \item \textbf{ARC} \cite{xumulti}: This dataset  consists of 7775 science multiple choice exam questions with answer options and 406 hierarchical labels. The average number of words per question in the dataset is 20.5. The number of train, validation and test samples are 5597, 778 and 1400 respectively.

     \item \textbf{Learning Objectives}: This dataset consists of 417 learning objectives collected from the \textit{"What you learnt"} section in class 8,9 and 10 science textbooks (K-12 system). The corresponding learning taxonomy was extracted from the "Table of contents" of the textbooks. 

 \end{itemize}

In our experiments we concatenate the question and the answer and it is considered as the input to the model (BERT), and the hierarchical taxonomy is considered as the label. Though BERT model has a context limit of 512 tokens, the length of each question-answer pair is within this range.

\subsection{Analysis of representation methods for encoding the hierarchical labels}
In this section, we briefly provide an analysis of different vector representation methods for projecting the hierarchical labels (learning taxonomy) to a continuous vector space. We embed the hierarchical labels using sentence representations methods like Sent2Vec \cite{pagliardini2017unsupervised} and Sentence-BERT \cite{reimers-gurevych-2019-sentence}.
             \begin{table}
       \small
 \centering
\caption{Comparison of different representation methods for hierarchical labels}\label{tab1}
\begin{tabular}{p{4.7cm}|p{3cm}|p{3cm}|c}
 Method & Label1 (L1)& Label2 (L2) & cos(L1, L2)\\
\hline\hline
Sentence-BERT &  science $\xrightarrow{}$ physics $\xrightarrow{}$ electricity &  science $\xrightarrow{}$ chemistry $\xrightarrow{}$ acids & 0.3072\\ 
Sent2vec & science $\xrightarrow{}$ physics $\xrightarrow{}$ electricity &  science $\xrightarrow{}$ chemistry $\xrightarrow{}$ acids & 0.6242\\ 
GloVe & science $\xrightarrow{}$ physics $\xrightarrow{}$ electricity &  science $\xrightarrow{}$ chemistry $\xrightarrow{}$ acids & 0.6632\\ 

\hline

\end{tabular}
\end{table}Additionally, we also average the word embeddings of individual terms in the hierarchical label using Glove to represent the label. We then compute the cosine similarity between the vectors of two different labels, and the results are as shown in Table 2. From Table 2, we observe that though "science $\xrightarrow{}$ physics $\xrightarrow{}$ electricity" and "science $\xrightarrow{}$ chemistry $\xrightarrow{}$ acids" are different, the representations obtained by averaging Glove embeddings output a high similarity score. This may be due to the loss of information by averaging word vectors. Additionally here, the context of words like electricity is not taken into account when encoding the word physics. Additionally, "physics" and "chemistry" are co-hyponyms which may result in their vectors being close in the continuous vector space. We also observe that Sent2Vec is also unable to capture the semantics of the labels as it gives a similar high cosine similarity score. However, we observe that the vectors obtained using Sentence-BERT are not very similar, as indicated by the cosine similarity score. This indicates that Sentence-BERT is able to produce semantically meaningful sentence representations for the hierarchical labels. We also observe that Sentence-BERT outputs high similarity scores for semantically related hierarchical labels. Since this analysis is not exhaustive, we also provide a detailed comparison of methods using different vector representation methods in Section 5.

\subsection{Methods and Experimental setup}
 We compare TagRec with flat multi-class classification methods and other state of the art methods. In TagRec, the labels are represented using transformer based sentence representation methods like Sentence-BERT (Sent\_BERT) \cite{reimers-gurevych-2019-sentence} or Universal Sentence Encoder \cite{cer-etal-2018-universal}.
The methods we compare against are:
\let\labelitemi\labelitemii
\begin{itemize}
  \setlength\itemsep{0.5em}

    \item \textbf{BERT+Sent2Vec} : In this method the training and testing phases are similar to TagRec. The labels representations are obtained using Sent2vec \cite{pagliardini2017unsupervised} instead of USE or Sent\_BERT.
    \item \textbf{BERT+GloVE} : In this method, the labels are represented as the average of the word embeddings of their constituent words. The word embeddings are obtained from GloVe.
    \setlength{\parindent}{0pt}
    \[V(label) = mean((Gl(subject),Gl(chapter),Gl(topic)))\]
    where, $V(label)$ denotes vector representation of the label, $Gl$ denotes GloVe pre-trained model. The training and testing phases are same as TagRec.

    \item \textbf{Twin BERT}:  This method is adapted from Twin BERT \cite{twinbert}. In this method, instead of using pre-trained sentence representation methods , we fine-tune a pre-trained BERT model to compute the label representations. The label representations correspond to the last layer hidden state of the first token. The first token is denoted as [CLS] in BERT, which is considered as the aggregate sequence representation. The BERT model that computes representations for the input and the BERT model for computing the label representations are fine-tuned simultaneously. 

    \item \textbf{BERT multi-class}  (label relation)  \cite{xumulti}: In this method, we fine-tune a pre-trained BERT model to classify the input question-answer pairs to one of the labels. Here the labels are encoded using label encoder, and hence this is a flat \textbf{multi-class classification} method. At inference time, we compute the representations for the question-answer pairs and labels using the fine-tuned model. Then the labels are ranked according to the cosine similarity scores computed between the input text embeddings and the label embeddings.

    \item \textbf{BERT multi-class} (prototypical embeddings) \cite{snell2017prototypical}: To provide a fair comparison with TagRec, we propose another baseline that considers the similarity between samples rather than the samples and the label. A BERT model is fine-tuned in a flat multi-class classification setting similar to the previous baseline. Then for each class, we compute a prototype, which is the mean of the embeddings of randomly chosen samples for each class from the training set. The embedding for each chosen sample is computed as the concatenation of the [CLS] token of the last 4 layers of the fine-tuned BERT model. We observe that this combination provides the best result for this baseline. After the prototypes are formed for each class, at inference time, we obtain the embeddings for each test sample in the same way and compute cosine similarity with the prototype embeddings for each class. Then the classes are ranked using the cosine similarity and top-k classes are returned.
    
    \item \textbf{Pretrained Sent\_BERT}: We implement a simple baseline where the vector representations of the input texts and the labels are obtained using a pre-trained Sentence-BERT model. There is no training involved in this baseline. For each input top closest matching labels are retrieved according to cosine similarity.
        \end{itemize}
  \begin{table}
 \centering
\caption{Performance comparison of TagRec with variants and baselines, $\dagger$ indicates TagRec's significant improvement at 0.001 level using \textit{t-test}}\label{tab1}
\begin{tabular}{p{1.1cm}|p{7.2cm}|p{1cm}|p{1cm}|p{1cm}|p{1cm}}
Dataset & Method &  R@5 & R@10& R@15 & R@20\\
\hline\hline
QC-Science & TagRec(BERT+USE) (proposed method) & \bf 0.86 &0.92 & \bf0.95 & 0.96\\ 
&TagRec(BERT+Sent\_BERT) (proposed method) & 0.85$\dagger$ & \bf 0.93$\dagger$ & \bf 0.95$\dagger$ & \bf 0.97$\dagger$\\ \cline{2-6}
&BERT+sent2vec & 0.79 &0.89 & 0.93 & 0.95\\ 
&Twin BERT \cite{twinbert} & 0.72 &0.86 & 0.91 & 0.94\\ 
&BERT+GloVe & 0.76 &0.87 & 0.92 & 0.94\\ \cline{2-6}
&BERT classification (label relation) \cite{xumulti} \ & 0.39 &0.50 & 0.57 & 0.63 \\ 
&BERT classification (prototypical embeddings) \cite{snell2017prototypical} & 0.83 &0.91 & 0.93 & 0.95 \\ \cline{2-6}
& Pretrained Sent\_BERT  & 0.30 &0.40 & 0.47 & 0.52 \\ 
\hline
ARC & TagRec(BERT+USE) (proposed method) & \bf 0.67$\dagger$ & \bf0.81$\dagger$ & \bf0.86$\dagger$ & \bf0.89$\dagger$\\ 
&TagRec(BERT+Sent\_BERT) (proposed method) & 0.65 &  0.77 &  0.84 &  0.88\\ \cline{2-6}
&BERT+sent2vec & 0.55 &0.72 & 0.81 & 0.87\\ 
&Twin BERT \cite{twinbert} & 0.46 &0.63 & 0.72 & 0.78\\ 
&BERT+GloVe & 0.56 &0.73 & 0.82 & 0.86\\ \cline{2-6}
&BERT classification (label relation) \cite{xumulti} \ & 0.27 &0.37 & 0.42 & 0.49 \\ 
&BERT classification (prototypical embeddings) \cite{snell2017prototypical} & 0.64 &0.75 & 0.80 & 0.83 \\ \cline{2-6}
& Pretrained Sent\_BERT  & 0.31 &0.46 & 0.54 & 0.59 \\ 
\hline
\end{tabular}
\end{table}


All the BERT models were fine-tuned for 30 epochs (with early stopping) with the ADAM optimizer, with learning rate of 2e-5 \cite{BERT} and epsilon which is a hyperparameter to avoid division by zero errors is set to 1e-8.
The random seed was set to a value of 42. The margin parameter in the hinge rank loss was set to a value of 0.1. All the implementations were done in Pytorch. The huggingface library \cite{wolf-etal-2020-transformers} was used to fine-tune pre-trained BERT models.

Our code and datasets are publicly available at  \url{https://bit.ly/3jQpzEv}

\section{Results and Discussion}
\setlength{\parindent}{0pt}

The performance comparison of the methods described in the previous section is shown in Table 3. We use the Recall@k metric, which is a common metric for ranked retrieval tasks. From the results, we observe that the proposed method TagRec (BERT+USE and BERT+Sent\_BERT) outperforms flat multi-class classification based baselines and other state of the art methods. We observe that representing the labels with transformer based sentence embedding methods perform the best. This is evident from the table as TagRec(BERT+USE) and TagRec(BERT+Sent\_BERT) outperform BERT+Sent2Vec and BERT+GloVe methods. This is because Universal Sentence Encoder (USE) and Sentence-BERT use self-attention to produce better representations. 
\begin{table}[h!]
\small
\caption{Examples demonstrating the performance for unseen labels at test time.}
\label{tab1}
\begin{tabular}{p{3.7cm}|p{2.9cm}|p{3.1cm}|p{2.5cm}}
Question text & Ground truth  & Top 2 predictions & Method\\ \hline \hline
A boy can see his face when he looks into a calm pond. Which physical property of the pond makes this happen? (A) flexibility (B) reflectiveness (C) temperature (D) volume& \multirow{5}{3cm}{matter$\xrightarrow{}$properties  of material$\xrightarrow{}$reflect} & \multirow{5}{3.2cm} {matter$\xrightarrow{}$properties of material$\xrightarrow{}$flex and}  \multirow{9}{3cm}{\textbf{matter$\xrightarrow{}$properties of material$\xrightarrow{}$reflect}} & \multirow{7}{2cm}{TagRec (BERT+USE) }
\\\cline{3-4}

 &  & matter$\xrightarrow{}$properties of objects$\xrightarrow{}$mass  \\ & & and \\& & matter$\xrightarrow{}$properties of objects$\xrightarrow{}$density & Twin BERT \cite{twinbert} \\\cline{3-4}
  &  & matter$\xrightarrow{}$states$\xrightarrow{}$solid and \\ & & matter$\xrightarrow{}$properties of material$\xrightarrow{}$density & BERT + GloVe \\\cline{3-4}
&  &   matter$\xrightarrow{}$properties of material$\xrightarrow{}$specific heat and \\ & & matter$\xrightarrow{}$properties \ of material& BERT+sent2vec \\ \hline
Which object best reflects light? (A) gray door (B) white floor (C) black sweater (D) brown carpet & \multirow{1}{3cm}{matter$\xrightarrow{}$ properties  of material$\xrightarrow{}$reflect} & \multirow{1}{3cm} {energy$\xrightarrow{}$light$\xrightarrow{}$reflect and}  \multirow{4}{3.2cm}{\textbf{matter$\xrightarrow{}$properties of material$\xrightarrow{}$reflect}} & \multirow{5}{2cm}{TagRec (BERT+USE)} 
\\ \cline{3-4}

 &  & energy$\xrightarrow{}$thermal$\xrightarrow{}$\\ & & radiation and \\ & & energy$\xrightarrow{}$light$\xrightarrow{}$generic properties & Twin BERT \cite{twinbert} \\\cline{3-4}
 &  & energy$\xrightarrow{}$light and \\ & & energy$\xrightarrow{}$light$\xrightarrow{}$refract & BERT + GloVe \\\cline{3-4}
&  &   energy$\xrightarrow{}$light$\xrightarrow{}$reflect and \\ & & energy$\xrightarrow{}$light$\xrightarrow{}$refract& BERT+sent2vec \\ \hline
\end{tabular}
\end{table}
  This reinforces the hypothesis that averaging the word vectors to represent the labels does not preserve the required semantic relatedness between labels. The Twin BERT architecture does not perform well when compared with TagRec. This is because the label representations obtained through fine-tuned BERT may not preserve the semantic relatedness than the label representations obtained from pre-trained sentence embedding models. 

 \begin{table}
 \small
 \centering
\caption{Performance comparison for learning objective categorization}\label{tab1}
\begin{tabular}{p{7.2cm}|p{1cm}|p{1cm}}
 Method & R@1& R@2\\
\hline\hline
 TagRec(BERT+USE) (proposed method) &  0.69 &0.85\\ 
TagRec(BERT+Sent\_BERT) (proposed method) & \bf 0.77 & \bf 0.91 \\ 
BERT+sent2vec & 0.49 &0.64\\ 
Twin BERT \cite{twinbert} & 0.54 &0.79\\ 
BERT+GloVe & 0.62 &0.84\\ 
BERT classification (label relation) \cite{xumulti} \ & 0.46 &0.59\\ 
BERT classification (prototypical embeddings) \cite{snell2017prototypical} & 0.60 &0.76 \\ 
 Pretrained Sent\_BERT  & 0.39 &0.54  \\ 
\hline

\end{tabular}
\end{table}
Also both the Sentence-BERT and the Universal Sentence Encoder models are trained on semantic text similarity (STS) tasks thereby rendering them the ideal candidates for retrieval based tasks. Finally we observe that the TagRec method outperforms the flat classification based baselines confirming the hypothesis that the representations learnt by aligning the input text and label representations provide better performance. This is pivotal to the task of question-answer pair categorization as the technical terms in the short input text are strongly correlated with the words in the label. The first baseline (BERT label relation) performs poorly as it has not been explicitly trained to minimize the distance between the input and label representations. This implies that the representations learnt through flat classification has no notion of label similarity. But the prototypical embeddings based baseline performs better as the classification is done based on similarity between train and test sample representations. However this baseline also has no notion of label similarity. Hence does not perform well when compared to our proposed method, TagRec. We also observe that the simple baseline of performing semantic search using pretrained Sentence-BERT does not work well as the model is not fine-tuned to align the input and labels.

To further show the efficacy of our method, we perform statistical significance tests and observe that the predicted results are statistically significant. For instance, for Recall@20 we observe that the predicted outputs from TagRec are statistically significant (\textit{t-test}) with p-values \textbf{0.000218} and \textbf{0.000816} for \textit{QC-Science} and \textit{ARC} respectively.

\setlength{\parindent}{4ex}The proposed method TagRec was also able to adapt to new labels. For instance, two samples in the test set of the \textbf{ARC} dataset were tagged with \textit{"matter$\xrightarrow{}$properties of material$\xrightarrow{}$reflect"} unseen during the training phase as shown in Table 4. At test time, the label \textit{"matter$\xrightarrow{}$properties of material$\xrightarrow{}$reflect"} appeared in top 2 predictions output by the proposed method (TagRec (BERT + USE)) for the two samples. We also observe that for the method (TagRec (BERT + Sent\_BERT)) the label \textit{"matter$\xrightarrow{}$properties of material$\xrightarrow{}$reflect"} appears in its top 5 predictions. We observe that for other methods shown in Table 4 the correct label does not appear even in top 10 predictions. The top 2 predictions from other methods for the samples are shown in Table 4. We also make similar observations for the BERT classification (label relation) and BERT classification (prototypical embeddings) baselines. We do not show them in Table 4 owing to space constraints. The top 2 predictions from BERT classification (prototypical embeddings) baseline for example 1 in Table 4 are \textit{matter$\xrightarrow{}$properties of objects$\xrightarrow{}$temperature} and \textit{matter$\xrightarrow{}$properties of objects$\xrightarrow{}$shape}.

For example 2, in Table 4, the top 2 predictions from BERT classification (prototypical embeddings) are \textit{energy$\xrightarrow{}$light$\xrightarrow{}$reflect} and \textit{matter$\xrightarrow{}$properties of material$\xrightarrow{}$color}.

The top 2 predictions from BERT classification (label relation) baseline for example 1 in Table 4 are \textit{matter$\xrightarrow{}$properties of objects$\xrightarrow{}$ density} and \textit{matter$\xrightarrow{}$\\properties of material$\xrightarrow{}$density}. For example 2, in Table 4, the top 2 predictions from BERT classification (label relation) are \textit{energy$\xrightarrow{}$light$\xrightarrow{}$refract} and \textit{matter$\xrightarrow{}$properties of material$\xrightarrow{}$luster}. This confirms our hypothesis that the proposed method can adapt to new labels without re-training or change in the model architecture unlike existing methods.

We also demonstrate the performance of TagRec on unseen but related learning content like the learning objectives. Learning objectives convey the learning goals and can be linked to learning content through the learning taxonomy.

We obtain the predictions for the given learning objectives using the models trained on the $QC-Science$ dataset. We do not fine-tune them on the given learning objectives dataset and directly use them as test set to obtain predictions. The results of the learning objective categorization task are shown in Table 5. We show the recall at top 1 and top 2 predictions as the best results were obtained in top 2 predictions. We observe that the proposed method TagRec outperforms other methods. Particularly TagRec (BERT + Sent\_BERT) which uses Sentence-BERT to represent the hierarchical labels gives the best performance. This demonstrates that the proposed method is able to generalize to unseen but related learning content without any re-training.
\section{Conclusion}
\setlength{\parindent}{0pt}
In this paper, we proposed a new method for learning to suggest hierarchical taxonomy (label) for short questions. We demonstrated that the representations learnt using the proposed similarity based learning method is better than flat classification methods and other state of the art methods \cite{twinbert}. Our method can easily adapt to unseen labels without a change in the architecture unlike flat classification based methods. We also demonstrated that the trained model can be used to categorize any related learning content like learning objectives without any retraining. The proposed method can also be used for taxonomy tagging in the forums like Quora and other discussion forums. The questions in Quora have a character limit of 50 words, but the answers could be longer than the context limit of the BERT model. To handle such long sequence lengths, we plan to explore new methods like Longformer \cite{beltagy2020longformer}. Also in the future, we aim to explore the hyperbolic space to represent the hierarchical labels.
\bibliographystyle{splncs04}

\end{document}